\documentclass[11pt]{article}
\usepackage{acl2015}
\usepackage{times}
\usepackage{latexsym}
\usepackage{amsmath}
\usepackage{url}
\usepackage{danudefs}
\usepackage{color}

\makeatletter
\newcommand{\@BIBLABEL}{\@emptybiblabel}
\newcommand{\@emptybiblabel}[1]{}
\makeatother
\usepackage{hyperref}

\usepackage{algorithmic}
\usepackage{algorithm}

\usepackage[font=small,skip=5pt]{caption}

\usepackage{graphicx}
\usepackage{epstopdf}

\usepackage{todonotes}

\title{Unsupervised Cross-Domain Word Representation Learning}

\author{Danushka Bollegala \ \ \ \ \ \ Takanori Maehara   \ \ \ \ \ \ Ken-ichi Kawarabayashi\\
\kern-5em {\normalsize \tt danushka.bollegala@} \ \ {\normalsize \tt maehara.takanori@} \ \ {\normalsize \tt \ \ \ \  k\_keniti@} \\
\kern-5em {\normalsize \tt \ \ \ liverpool.ac.uk} \ \ \ \ {\normalsize \tt \ shizuoka.ac.jp} \ \ {\normalsize \tt \ \ \ \  \ \ \ nii.ac.jp} \\
\ \ \ \ \  \ University of Liverpool \ \ \ \ Shizuoka University  \ \ \ \ \ National Institute of Informatics \\
JST, ERATO, Kawarabayashi Large Graph Project. 
}

\date{}

\begin{document}
\maketitle

\begin{abstract}
Meaning of a word varies from one domain to another. 
Despite this important domain dependence in word semantics, 
existing word representation learning methods are bound to a single domain.
Given a pair of \emph{source}-\emph{target} domains, 
we propose an unsupervised method for learning domain-specific word representations
that accurately capture the domain-specific aspects of word semantics.
First, we select a subset of frequent words that occur in both domains as \emph{pivots}.
Next, we optimize an objective function that enforces two constraints:
(a) for both source and target domain documents, 
pivots that appear in a document must accurately predict the co-occurring non-pivots, and
(b) word representations learnt for pivots must be similar in the two domains.
Moreover, we propose a method to perform domain adaptation using the learnt word representations.
Our proposed method significantly outperforms competitive baselines including the
state-of-the-art domain-insensitive word representations, and reports best sentiment classification
accuracies for all domain-pairs in a benchmark dataset.
\end{abstract}

\section{Introduction}
\label{sec:intro}

Learning semantic representations for words is a fundamental task in NLP
that is required in numerous higher-level NLP applications~\cite{Collobert:2011}.
Distributed word representations have gained much popularity lately because of their accuracy
as semantic representations for words~\cite{Milkov:2013,Pennington:EMNLP:2014}. 
However, the meaning of a word often varies from one domain to another.
For example, the phrase \emph{lightweight} is often used in a positive sentiment in the
portable electronics domain because a lightweight device is easier to carry around,
which is a positive attribute for a portable electronic device.
However, the same phrase has a negative sentiment assocition in the \emph{movie}
domain because movies that do not invoke deep thoughts in viewers are considered to be lightweight~\cite{Bollegala:ACL:2014}. 
However, existing word representation learning methods are agnostic to such domain-specific semantic variations
of words, and capture semantics of words only within a single domain.
To overcome this problem and capture domain-specific semantic orientations of words, 
we propose a method that learns separate distributed representations for each domain in which a word occurs.

Despite the successful applications of distributed word representation 
learning methods~\cite{Pennington:EMNLP:2014,Collobert:2011,Milkov:2013}
most existing approaches are limited to learning only a single representation for a given word~\cite{Reisinger:NAACL:2010}.
Although there have been some work on learning multiple \emph{prototype} 
representations~\cite{Huang:ACL:2012,neelakantan-EtAl:2014:EMNLP2014}
for a word considering its multiple senses, such methods do not consider the semantics of the domain
in which the word is being used. 

If we can learn separate representations for a word for each
domain in which it occurs, we can use the learnt representations for domain adaptation tasks
such as cross-domain sentiment classification~\cite{Bollegala:ACL:2011}, 
cross-domain POS tagging~\cite{schnabel-schutze:2013:IJCNLP},
cross-domain dependency parsing~\cite{McClosky:NAACL:2010}, and
domain adaptation of 
relation extractors~\cite{Bollegala:IJCAI:2013,Bollegala:TKDE:RA:2013,Bollegala:IJCAI:2011,Jiang:ACL:2007,Jiang:CIKM:2007}.

We introduce the \emph{cross-domain word representation learning} task, where given two domains,
 (referred to as the \emph{source} ($\cS$) and the \emph{target} ($\cT$))
the goal is to learn two separate representations $\vec{w}_\cS$ and $\vec{w}_\cT$ for a word
$w$ respectively from the source and the target domain that capture \emph{domain-specific} semantic variations of $w$. 
In this paper, we use the term \emph{domain} to represent a collection of documents related to
a particular topic such as user-reviews in Amazon for a product category (e.g. \textit{books}, \textit{dvds}, \textit{movies}, etc.). 
However, a domain in general can be a field of study (e.g. \textit{biology}, \textit{computer science}, \textit{law}, etc.)
or even an entire source of information (e.g. \textit{twitter}, \textit{blogs}, \textit{news articles}, etc.).
In particular, we do not assume the availability of any labeled data for learning word representations.

This problem setting is closely related to unsupervised domain adaptation~\cite{Blitzer:EMNLP:2006}, 
which has found numerous useful applications such as, sentiment classification and POS tagging.
For example, in unsupervised cross-domain sentiment classification~\cite{Blitzer:EMNLP:2006,Blitzer:ACL:2007}, 
we train a binary sentiment classifier using
positive and negative labeled user reviews in the source domain, and apply the trained classifier to predict
sentiment of the target domain's user reviews.
Although the distinction between the source and the target domains is not important for the word representation learning
step, it is important for the domain adaptation tasks in which we subsequently evaluate the learnt word representations.
Following prior work on domain adaptation~\cite{Blitzer:EMNLP:2006},
high-frequent features (unigrams/bigrams) common to both domains are referred to as \emph{domain-independent}
features or \emph{pivots}. 
In contrast, we use \emph{non-pivots} to refer to features that are specific to a single domain.

We propose an unsupervised cross-domain word representation learning method that jointly optimizes two criteria:
(a) given a document $d$ from the source or the target domain, we must accurately
predict the non-pivots that occur in $d$ using the pivots that occur in $d$, and
(b) the source and target domain representations we learn for pivots must be similar.
The main challenge in domain adaptation is \emph{feature mismatch}, where the features that we use for training
a classifier in the source domain do not necessarily occur in the target domain. 
Consequently, prior work on domain adaptation~\cite{Blitzer:EMNLP:2006,Pan:WWW:2010}
learn lower-dimensional mappings from non-pivots to pivots, thereby overcoming the feature mismatch problem.
Criteria (a) ensures that word representations for domain-specific non-pivots in each domain are related to
the word representations for domain-independent pivots. This relationship enables us to discover
pivots that are similar to target domain-specific non-pivots, thereby overcoming the feature mismatch problem.

On the other hand, criteria (b) captures the prior knowledge that high-frequent words 
common to two domains often represent domain-independent
semantics. For example, in sentiment classification, words such as \emph{excellent} or \emph{terrible} would express
similar sentiment about a product irrespective of the domain. 
However, if a pivot expresses different semantics in source and the target domains,
then it will be surrounded by dissimilar sets of non-pivots, and reflected in the first criteria.
Criteria (b) can also be seen as a regularization constraint imposed on word representations to prevent overfitting by
reducing the number of free parameters in the model. 

Our contributions in this paper can be summarized as follows.
\begin{itemize}
\item We propose a distributed word representation learning method that learns separate representations for a word
for each domain in which it occurs. 
To the best of our knowledge, ours is the first-ever \emph{domain-sensitive distributed} word representation learning method.
\item Given domain-specific word representations, we propose a method to learn a cross-domain sentiment classifier.

Although word representation learning methods have been used for various related tasks in NLP such as 
similarity measurement~\cite{Mikolov:NAACL:2013}, POS tagging~\cite{Collobert:2011}, 
dependency parsing~\cite{Socher:ICML:2011}, machine translation~\cite{Zou:EMNLP:2013}, 
sentiment classification~\cite{socher-EtAl:2011:EMNLP},
and semantic role labeling~\cite{roth-woodsend:2014:EMNLP2014}, to the best of our knowledge,
word representations methods have not yet been used for cross-domain sentiment classification.
\end{itemize}

Experimental results for cross-domain sentiment classification on a benchmark
dataset show that the word representations learnt using the proposed method 
statistically significantly outperform a state-of-the-art domain-insensitive word representation learning method~\cite{Pennington:EMNLP:2014},
and several competitive baselines. In particular, our proposed cross-domain word representation learning method
is not specific to a particular task such as sentiment classification, and in principle, can be in applied to a wide-range 
of domain adaptation tasks.
Despite this task-independent nature of the proposed method, it achieves the best sentiment classification accuracies
on all domain-pairs, reporting statistically comparable results to the
current state-of-the-art unsupervised cross-domain sentiment classification methods~\cite{Pan:WWW:2010,Blitzer:EMNLP:2006}.

\section{Related Work}
\label{sec:related}

Representing the semantics of a word using some algebraic structure such as a vector (more generally a tensor)
is a common first step in many NLP tasks~\cite{Turney:JAIR:2010}. By applying algebraic operations on the word representations,
we can perform numerous tasks in NLP, such as composing representations for larger textual units beyond individual words such
as phrases~\cite{Mitchell:ACL:2008}. Moreover, word representations are found to be useful for measuring semantic similarity,
and for solving proportional analogies~\cite{Mikolov:NAACL:2013}. 
Two main approaches for computing word representations can be identified
in prior work~\cite{baroni-dinu-kruszewski:2014:P14-1}: \emph{counting-based} and \emph{prediction-based}.

In counting-based approaches~\cite{Baroni:DM}, a word $w$ is represented by a vector $\vec{w}$ that contains other words that
co-occur with $w$ in a corpus. Numerous methods for selecting co-occurrence contexts
such as proximity or dependency relations have been proposed~\cite{Turney:JAIR:2010}. 
Despite the numerous successful applications of co-occurrence counting-based distributional word representations,
their high dimensionality and sparsity are often problematic in practice.
Consequently, further post-processing steps such as dimensionality reduction,
and feature selection are often required when using counting-based word representations.

On the other hand, prediction-based approaches first assign each word, for example, with a $d$-dimensional
real-vector, and learn the elements of those vectors by applying them in an auxiliary task such as language modeling,
where the goal is to predict the next word in a given sequence.
The dimensionality $d$ is fixed for all the words in the vocabulary, and,  
unlike counting-based word representations, is much smaller (e.g. $d \in [10, 1000]$ in practice) compared to the vocabulary size.
The neural network language model (NNLM)~\cite{Bengio:JMLR:2003}
uses a multi-layer feed-forward neural network to predict the next word in a sequence,
and uses backpropagation to update the word vectors such that the prediction error is minimized.

Although NNLMs learn word representations as a by-product, the main focus on language modeling is
to predict the next word in a sentence given the previous words, and not learning word representations that capture semantics.
Moreover, training multi-layer neural networks using large text corpora is time consuming.
To overcome those limitations, methods that specifically focus on learning word
representations that model word co-occurrences in large corpora
 have been proposed~\cite{Milkov:2013,Mnih:2013,Huang:ACL:2012,Pennington:EMNLP:2014}.
Unlike the NNLM, these methods use \emph{all} the words in a contextual window in the prediction task.
Methods that use one or no hidden layers are proposed to improve the scalability of the learning algorithms.
For example, the skip-gram model~\cite{Mikolov:NIPS:2013} predicts the words $c$ that appear in the local context of a word $w$, whereas
the continuous bag-of-words model (CBOW) predicts a word $w$ conditioned on all the words $c$ that appear in
$w$'s local context~\cite{Milkov:2013}.
Methods that use global co-occurrences in the entire corpus to learn word
representations have shown to outperform methods that use only local co-occurrences~\cite{Huang:ACL:2012,Pennington:EMNLP:2014}.
Overall, prediction-based methods have shown to outperform counting-based methods~\cite{baroni-dinu-kruszewski:2014:P14-1}.

Despite their impressive performance,  existing methods for word representation learning
do not consider the semantic variation of words across different domains.
However, as described in Section~\ref{sec:intro},
the meaning of a word vary from one domain to another, and must be considered.
To the best of our knowledge,
the only prior work studying the problem of word representation variation across domains is due to
Bollegala et al.~\shortcite{Bollegala:ACL:2014}. Given a source and a target domain,
they first select a set of pivots using pointwise mutual information, and create two distributional representations for each pivot using
their co-occurrence contexts in a particular domain. Next, a projection matrix from the source to the target domain feature spaces
is learnt using partial least squares regression. Finally, the learnt projection matrix is used to
find the nearest neighbors in the source domain for each target domain-specific features.
However, unlike our proposed method, their method \emph{does not} learn domain-specific word representations,
but simply uses co-occurrence counting when creating in-domain word representations.

Faralli et al. \shortcite{Faralli:EMNLP:2012} proposed a domain-driven word sense disambiguation (WSD) method where they
construct glossaries for several domain using a pattern-based bootstrapping technique.
This work demonstrates the importance of considering the domain specificity of word senses.
However, the focus of their work is not to learn representations for words or their senses in a domain, but to
construct glossaries. It would be an interesting future research direction to explore the possibility
of using such domain-specific glossaries for learning domain-specific word representations.

Neelakantan et al.~\shortcite{neelakantan-EtAl:2014:EMNLP2014} proposed a method that jointly performs WSD and word embedding learning, thereby
learning multiple embeddings per word type. In particular, the number of senses per word type is
automatically estimated. However, their method is limited to a single domain, and does not consider how the representations vary across domains. 
On the other hand, our proposed method learns a single representation for a particular
word for each domain in which it occurs.

Although in this paper we focus on the monolingual setting where source and target domains belong to the same language,
the related setting where learning representations for words that are translational pairs across languages
 has been studied~\cite{Moritz:ICLR:2014,Klementiev:COLING:2012,Gouws:ICML:2015}.
Such representations are particularly useful for cross-lingual information retrieval~\cite{Duc:WI:2010}.
It will be an interesting future research direction to extend our proposed method to learn such cross-lingual word representations.

\section{Cross-Domain Representation Learning}

We propose a method for learning word representations that are sensitive to the semantic variations of words
across domains. We call this problem \emph{cross-domain word representation learning}, and provide a definition
 in Section~\ref{sec:definition}. Next, in Section~\ref{sec:model}, given a set of pivots that occurs in both a source
and a target domain, we propose a method for learning cross-domain word representations.
We defer the discussion of pivot selection methods to Section~\ref{sec:pivots}.
In Section~\ref{sec:DA}, we propose a method for using the learnt word representations to train
a cross-domain sentiment classifier.

\subsection{Problem Definition}
\label{sec:definition}

Let us assume that we are given two sets of documents $\cD_\cS$ and $\cD_\cT$ respectively for a source ($\cS$)
and a target ($\cT$) domain. 
We do not consider the problem of retrieving documents for a domain, and assume such a collection of documents to be given.
Then, given a particular word $w$, we define cross-domain representation learning as the task of learning two separate representations
$\vec{w}_\cS$ and $\vec{w}_\cT$ capturing $w$'s semantics in respectively the source $\cS$ and the target $\cT$ domains.

Unlike in domain adaptation, where there is a clear distinction between the source (i.e. the domain on which we train)
vs. the target (i.e. the domain on which we test) domains, for representation learning purposes we do not make a distinction
between the two domains. 
In the \emph{unsupervised} setting of the cross-domain representation learning that we
study in this paper, we do not assume the availability of labeled data for any domain for the purpose of learning word representations.
As an extrinsic evaluation task, we apply the trained word representations
for classifying sentiment related to user-reviews (Section~\ref{sec:DA}).
However, for this evaluation task we require sentiment-labeled user-reviews from the
source domain. 

Decoupling of the word representation learning from any tasks in which those representations
are subsequently used, simplifies the problem as well as enables us to learn \emph{task-independent} word representations
with potential generic applicability.
Although we limit the discussion to a pair of domains for simplicity,
the proposed method can be easily extended to jointly learn word representations for more than two domains. 
In fact, prior work on cross-domain sentiment analysis show that
incorporating multiple source domains improves sentiment classification 
accuracy on a target domain~\cite{Bollegala:ACL:2011,Glorot:ICML:2011}.

\subsection{Proposed Method}
\label{sec:model}

To describe our proposed method, let us denote a pivot and a non-pivot feature respectively by $c$ and $w$. 
Our proposed method does not depend on a specific pivot selection method,
and can be used with all previously proposed methods for selecting pivots as explained later in Section~\ref{sec:pivots}.
A pivot $c$ is represented in the source and target domains respectively
by vectors $\vec{c}_\cS \in \R^n$ and $\vec{c}_\cT \in \R^n$.
Likewise, a source specific non-pivot $w$ is represented by $\vec{w}_\cS$ in the source domain,
whereas a target specific non-pivot $w$ is represented by $\vec{w}_\cT$ in the target domain.
By definition, a non-pivot occurs only in a single domain. For notational convenience we use $w$ to
denote non-pivots in both domains when the domain is clear from the context.
We use $\cC_\cS$, $\cW_\cS, \cC_\cT$, and $\cW_\cT$ to denote the sets of word representation vectors
respectively for the source pivots, source non-pivots, target pivots, and target non-pivots.

Let us denote the set of documents in the source and the target domains respectively by $\cD_\cS$ and $\cD_\cT$.
Following the bag-of-features model, we assume that a document $D$ is represented by the set of pivots and non-pivots that occur in $D$ 
($w \in d$ and $c \in d$).
We consider the co-occurrences of a pivot $c$ and a non-pivot $w$ within a fixed-size contextual window in a document.
Following prior work on representation learning~\cite{Milkov:2013}, in our experiments, 
we set the window size to $10$ tokens, without crossing sentence boundaries.
The notation $(c, w) \in d$ denotes the co-occurrence of a pivot $c$ and a non-pivot $w$ in a document $d$. 

We learn domain-specific word representations by maximizing the
prediction accuracy of the non-pivots $w$ that occur in the local context of a pivot $c$. 
The hinge loss, $L(\cC_\cS, \cW_\cS)$, associated with predicting a non-pivot $w$ in a source document $d \in \cD_\cS$
that co-occurs with pivots $c$ is given by:
\begin{equation}
\small
\label{eq:source-loss}
\sum_{d \in \cD_\cS} \sum_{(c, w) \in d} \sum_{w^*\!\by\!p(w)} \max  \left(0, 1 -{ \vec{c}_\cS}\T \vec{w}_\cS + {\vec{c}_\cS}\T \vec{w}^{*}_\cS \right) 
\end{equation}
Here, $\vec{w}^{*}_\cS$ is the source domain representation of a non-pivot $w^*$ that \emph{does not occur} in $d$.
The loss function given by Eq.~\ref{eq:source-loss} requires that a non-pivot $w$ that co-occurs with a pivot $c$ 
in the document $d$ is assigned a higher ranking score as measured by the inner-product between $\vec{c}_\cS$
and $\vec{w}_\cS$ than a non-pivot $w^*$ that does not occur in $d$. We randomly sample $k$
non-pivots from the set of all source domain non-pivots that do not occur in $d$ as $w^*$. 

Specifically, we use the marginal distribution of non-pivots $p(w)$, estimated from the corpus counts, as the sampling distribution. 
We raise $p(w)$ to the $3/4$-th power as proposed by Mikolov et al.~\shortcite{Milkov:2013},
and normalize it to unit probability mass prior to sampling $k$ non-pivots $w^*$ per each co-occurrence of $(c,w) \in d$.
Because non-occurring non-pivots $w^*$ are randomly sampled, prior work on noise contrastive estimation
has found that it requires more negative samples than positive samples to accurately learn a prediction model~\cite{Mnih:2013}.
We experimentally found $k = 5$ to be an acceptable trade-off between the prediction accuracy and the number of training instances.

Likewise, the loss function $L(\cC_\cT, \cW_\cT)$ for predicting non-pivots using pivots in the target domain is given by:
\begin{equation}
\small
\label{eq:target-loss}
\sum_{d \in \cD_\cT} \sum_{(c,w) \in d} \sum_{w^*\!\by\!p(w)} \max  \left(0, 1 -{ \vec{c}_\cT}\T \vec{w}_\cT + {\vec{c}_\cT}\T \vec{w}^{*}_\cT \right)
\end{equation}
Here, $w^*$ denotes target domain non-pivots that \emph{do not occur} in $d$, and are randomly sampled from $p(w)$ following the same procedure as in the source domain.

The source and target loss functions given respectively by Eqs.~\ref{eq:source-loss} and \ref{eq:target-loss}
can be used on their own to independently learn source and target domain word representations. 
However, by definition, pivots are common to both domains. 
We use this property to relate the source and target word representations via a \emph{pivot-regularizer}, $R(\cC_\cS, \cC_\cT)$, defined as:
\begin{equation}
\label{eq:pivot-regularizer}
R(\cC_\cS, \cC_\cT) = \frac{1}{2}\sum_{i=1}^{K} {||\vec{c}^{(i)}_\cS - \vec{c}^{(i)}_\cT||}^2 
\end{equation}
Here, $||\vec{x}||$ represents the $l_2$ norm of a vector $\vec{x}$, and $c^{(i)}$ is the $i$-th pivot in a total collection of $K$ pivots.
Word representations for non-pivots in the source and target domains are linked via the pivot regularizer because,
the non-pivots in each domain are predicted using the word representations for the pivots in each domain, 
which in turn are regularized by Eq.~\ref{eq:pivot-regularizer}.
The overall objective function, $L(\cC_\cS, \cW_\cS, \cC_\cT, \cW_\cT)$, we minimize is the sum\footnote{Weighting the source and
target loss functions by the respective dataset sizes did not result in any significant increase in performance. We believe that this is 
because the benchmark dataset contains approximately equal numbers of documents for each domain.} of the source and target loss functions, regularized via Eq.~\ref{eq:pivot-regularizer} with coefficient $\lambda$, and is given by:
\begin{equation}
\label{eq:overall}
L(\cC_\cS, \cW_\cS,) + L(\cC_\cT, \cW_\cT) + \lambda R(\cC_\cS, \cC_\cT) 
\end{equation}

\subsection{Training}
\label{sec:train}

Word representations of pivots $c$ and non-pivots $w$ 
in the source ($\vec{c}_\cS$, $\vec{w}_\cS$) and the target ($\vec{c}_\cT$, $\vec{w}_\cT$) domains are
parameters to be learnt in the proposed method.
To derive parameter updates, we compute the gradients of the overall loss function in Eq.~\ref{eq:overall}
w.r.t. to each parameter as follows:

{\small
\begin{align}
&\frac{\partial L}{\partial \vec{w}_\cS} =
\begin{cases}
0 & \text{if \ \ } \vec{c}_\cS\T (\vec{w}_\cS - \vec{w}^*_\cS) \geq 1 \\
-\vec{c}_\cS & \text{otherwise}
\end{cases} \\
&\frac{\partial L}{\partial \vec{w}^*_\cS} =
\begin{cases}
0 & \text{if \ \ } \vec{c}_\cS\T (\vec{w}_\cS - \vec{w}^*_\cS) \geq 1 \\
\vec{c}_\cS & \text{otheriwse}
\end{cases} \\
&\frac{\partial L}{\partial \vec{w}_\cT} =
\begin{cases}
0 & \text{ if \ \ } \vec{c}_\cT\T (\vec{w}_\cT - \vec{w}^*_\cT) \geq 1 \\
-\vec{c}_\cT & \text{otherwise}
\end{cases} \\
& \frac{\partial L}{\partial \vec{w}^*_\cT} =
\begin{cases}
0 & \text{ if \ \ } \vec{c}_\cT\T (\vec{w}_\cT - \vec{w}^*_\cT) \geq 1 \\
\vec{c}_\cT & \text{otherwise}
\end{cases} \\
&\frac{\partial L}{\partial \vec{c}_\cS} =
\begin{cases}
\lambda (\vec{c}_\cS - \vec{c}_\cT) & \text{\!\!\!\!\!\!\!\!\!\!if \ \ } \vec{c}_\cS\T (\vec{w}_\cS - \vec{w}^*_\cS) \geq 1 \\
\vec{w}^*_\cS - \vec{w}_\cS + \lambda(\vec{c}_\cS - \vec{c}_\cT)  & \text{otherwise}
\end{cases} \\
&\frac{\partial L}{\partial \vec{c}_\cT} =
\begin{cases}
\lambda(\vec{c}_\cT - \vec{c}_\cS) &  \text{\!\!\!\!\!\!\!\!\!\!if \ \ } \vec{c}_\cT\T (\vec{w}_\cT - \vec{w}^*_\cT) \geq 1 \\
\vec{w}^*_\cT- \vec{w}_\cT + \lambda (\vec{c}_\cT - \vec{c}_\cS)  & \text{otherwise}
\end{cases} 
\end{align}
}%
Here, for simplicity, we drop the arguments inside the loss function and write it as $L$.
We use mini batch stochastic gradient descent with a batch size of $50$ instances.
AdaGrad~\cite{Duchi:JMLR:2011} is used to schedule the learning rate.
All word representations are initialized with $n$ dimensional random vectors sampled from 
a zero mean and unit variance Gaussian. Although the objective in Eq.~\ref{eq:overall} is not jointly convex in 
all four representations, it is convex w.r.t. the representation of a particular feature (pivot or non-pivot)
when the representations for all the other features are held fixed. In our experiments, the training 
converged in all cases with less than $100$ epochs over the dataset.

The rank-based predictive hinge loss (Eq.~\ref{eq:source-loss}) is inspired by the prior work on word representation learning
for a single domain~\cite{Collobert:2011}.
However, unlike the multilayer neural network  in \newcite{Collobert:2011}, the proposed method uses
a computationally efficient single layer to reduce the number of parameters that must be learnt, thereby scaling to large datasets.
Similar to the skip-gram model~\cite{Milkov:2013}, the proposed method predicts occurrences of contexts (non-pivots) $w$ 
within a fixed-size contextual window of a target word (pivot) $c$.

Scoring the co-occurrences of two words $c$ and $w$ by the bilinear form given by the 
inner-product is similar to prior work on domain-insensitive word-representation 
learning~\cite{Mnih:HLBL:NIPS:2008,Milkov:2013}.
However, unlike those methods that use the softmax function to convert inner-products to probabilities,
we directly use the inner-products without any further transformations, thereby avoiding
computationally expensive distribution normalizations over the entire vocabulary.

\subsection{Pivot Selection}
\label{sec:pivots}

Given two sets of documents $\cD_\cS$, $\cD_\cT$ respectively for the source and the target domains,
we use the following procedure to select pivots and non-pivots.
First, we tokenize and lemmatize each document using the 
Stanford CoreNLP toolkit\footnote{\url{http://nlp.stanford.edu/software/corenlp.shtml}}.
Next, we extract unigrams and bigrams as features for representing a document. 
We remove features listed as stop words using a standard stop words list.
Stop word removal increases the effective co-occurrence window size for a pivot.
Finally, we remove features that occur less than $50$ times in the entire set of documents.

Several methods have been proposed in the prior work on domain adaptation for selecting
a set of pivots from a given pair of domains such as the minimum frequency of occurrence of a feature
in the two domains, mutual information (MI), and the entropy of the feature distribution over the documents~\cite{Pan:WWW:2010}.
In our preliminary experiments, we discovered that a normalized version of the PMI (NPMI)~\cite{Bouma:2009}
to work consistently well for selecting pivots from different pairs of domains.
NPMI between two features $x$ and $y$ is given by:
\begin{equation}
\label{eq:NPMI}
\mathrm{NPMI}(x, y) = \log\left(\frac{p(x,y)}{p(x)p(y)}\right) \frac{1}{-\log(p(x,y))} 
\end{equation}
Here, the joint probability $p(x,y)$, and the marginal probabilities $p(x)$ and $p(y)$ are estimated
using the number of co-occurrences of $x$ and $y$ in the sentences in the documents.
Eq.~\ref{eq:NPMI} normalizes both the upper and lower bounds of the PMI.

We measure the appropriateness of a feature as a pivot according to the score given by:
\begin{equation}
\label{eq:pivot-score}
\mathrm{score}(x) = \min \left( \mathrm{NPMI}(x, \cS), \mathrm{NPMI}(x, \cT) \right) .
\end{equation}
We rank features that are common to both domains in the descending order of
their scores as given by Eq.~\ref{eq:pivot-score}, and select the top $N_\cP$ features as pivots.
We rank features $x$ that occur only in the source domain by $\mathrm{NPMI}(x, \cS)$,
and select the top ranked $N_\cS$ features as source-specific non-pivots.
Likewise, we rank the features $x$ that occur only in the target domain by $\mathrm{NPMI}(x, \cT)$,
and select the top ranked $N_\cT$ features as target-specific non-pivots.

The pivot selection criterion described here differs from that of
 Blitzer et al.~\shortcite{Blitzer:EMNLP:2006,Blitzer:ACL:2007},
where pivots are defined as features that behave similarly both in the source and the target domains.
They compute the mutual information between a feature (i.e. unigrams or bigrams) and the sentiment labels
using source domain labeled reviews. This method is useful when selecting pivots that are closely
associated with positive or negative sentiment in the source domain. However, in unsupervised domain adaptation
we do not have labeled data for the target domain. Therefore, the pivots selected using this approach are not
guaranteed to demonstrate the same sentiment in the target domain as in the source domain.
On the other hand, the pivot selection method proposed in this paper focuses on identifying a subset of
features that are closely associated with both domains. 

It is noteworthy that our proposed cross-domain word representation learning method (Section~\ref{sec:model})
\emph{does not} assume any specific pivot/non-pivot selection method.
Therefore, in principle, our proposed word representation learning method could be used with any of the
previously proposed pivot selection methods. 
We defer a comprehensive evaluation of possible combinations of pivot selection methods and their effect on
the proposed word representation learning method to future work.

\subsection{Cross-Domain Sentiment Classification}
\label{sec:DA}

As a concrete application of cross-domain word representations, we describe a method for learning a cross-domain sentiment
classifier using the word representations learnt by the proposed method. 
Existing word representation learning methods that learn from only a single domain are typically evaluated for their
accuracy in measuring semantic similarity between words, or by solving word analogy problems.
Unfortunately, such gold standard datasets capturing cross-domain semantic variations of words are unavailable.
Therefore, by applying the learnt word representations in a cross-domain sentiment classification task,
we can conduct an indirect extrinsic evaluation. 

The train data available for unsupervised cross-domain sentiment classification
consists of unlabeled data for both the source and the target domains
as well as labeled data for the source domain. We train a binary sentiment classifier using those train data,
and apply it to classify sentiment of the target test data. 

Unsupervised cross-domain sentiment classification is challenging due to two reasons: \emph{feature-mismatch},
and \emph{semantic variation}.
First, the sets of features that occur in source and target domain documents are different.
Therefore, a sentiment classifier trained using source domain labeled data is likely to encounter unseen features during
test time. We refer to this as the feature-mismatch problem. Second, some of the features that occur in both domains
will have different sentiments associated with them (e.g. \emph{lightweight}). Therefore, a sentiment classifier trained
using source domain labeled data is likely to incorrectly predict similar sentiment (as in the source) for such features.
We call this the semantic variation problem. Next, we propose a method to overcome both problems using cross-domain word representations.

Let us assume that we are given a set $\{(\vec{x}^{(i)}_{\cS}, y^{(i)})\}_{i=1}^{n}$ of $n$ labeled reviews $\vec{x}^{(i)}_{\cS}$ 
 for the source domain $\cS$. For simplicity, let us consider binary sentiment classification where each review $\vec{x}^{(i)}$ is 
 labeled either as positive (i.e.\ $y^{(i)} = 1$) or negative (i.e.\ $y^{(i)} = -1$). 
 Our cross-domain binary sentiment classification method can be easily extended to multi-class classification.
 First, we lemmatize each word in a source domain labeled review $\vec{x}^{(i)}_{\cS}$, 
 and extract unigrams and bigrams as features to represent $\vec{x}^{(i)}_{\cS}$ by a binary-valued feature vector.
 Next, we train a binary linear classifier, $\vec{\theta}$, using those feature vectors.
 Any binary classification algorithm can be used for this purpose.
We use $\vec{\theta}(z)$ to denote the weight learnt by the classifier for a feature $z$.
 In our experiments, we used $l_2$ regularized logistic regression. 
 
At test time, we represent a test target review by a binary-valued vector $\vec{h}$ using a the set of unigrams and bigrams 
extracted from that review. Then, the activation score, $\psi(\vec{h})$, of $\vec{h}$ is defined by:
\begin{equation}
\small
\psi(\vec{h}) = \sum_{c \in \vec{h}} \sum_{c' \in \vec{\theta}} \vec{\theta}(c') f(\vec{c}'_\cS, \vec{c}_\cS) +
						\sum_{w \in \vec{h}} \sum_{w' \in \vec{\theta}} \vec{\theta}(w') f(\vec{w}'_\cS, \vec{w}_\cT)
\label{eq:score}
\end{equation}
Here, $f$ is a similarity measure between two vectors. If $\psi(\vec{h}) > 0$, we classify $\vec{h}$ as positive,
and negative otherwise. Eq.~\ref{eq:score} measures the similarity between each feature in $\vec{h}$ against the
features in the classification model $\vec{\theta}$.
For pivots $c \in \vec{h}$, we use the the source domain representations to measure
similarity, whereas for the (target-specific) non-pivots $w \in \vec{h}$, we use their target domain representations.
We experimented with several popular similarity measures for $f$ and found cosine similarity to perform consistently well.
We can interpret Eq.~\ref{eq:score} as a method for \emph{expanding} a test target document using
nearest neighbor features from the source domain labeled data.
It is analogous to query expansion used in information retrieval to improve document recall~\cite{Fang:ACL:2008}.
Alternatively, Eq.~\ref{eq:score} can be seen as a linearly-weighted additive kernel function over two feature spaces.

\section{Experiments and Results}
\label{sec:exp}

For train and evaluation purposes, we use the Amazon product reviews collected by \newcite{Blitzer:ACL:2007} 
for the four product categories:
books (\textbf{B}), DVDs (\textbf{D}), electronic items (\textbf{E}), and kitchen appliances (\textbf{K}).
There are $1000$ positive and $1000$ negative sentiment labeled reviews for each domain.
Moreover, each domain has on average $17,547$ unlabeled reviews.
We use the standard split of $800$ positive and $800$ negative labeled reviews from each domain
as training data, and the rest (200+200) for testing. 
For validation purposes we use \textit{movie} (source) and
\textit{computer} (target) domains, which were also collected by \newcite{Blitzer:ACL:2007}, but not part of the train/test domains.

Experiments conducted using this validation dataset revealed that the performance of the proposed method
is relatively insensitive to the value of the regularization parameter $\lambda \in [10^{-3}, 10^{3}]$.
For the non-pivot prediction task we generate positive and negative instances using the procedure described in Section~\ref{sec:model}.
As a typical example, we have $88,494$ train instances from the books source domain and $141,756$ train instances
from the target domain (1:5 ratio between positive and negative instances in each domain).
The number of pivots and non-pivots are set to $N_\cP = N_\cS = N_\cT = 500$.

\begin{figure*}[t]
\centering
\includegraphics[width=17cm]{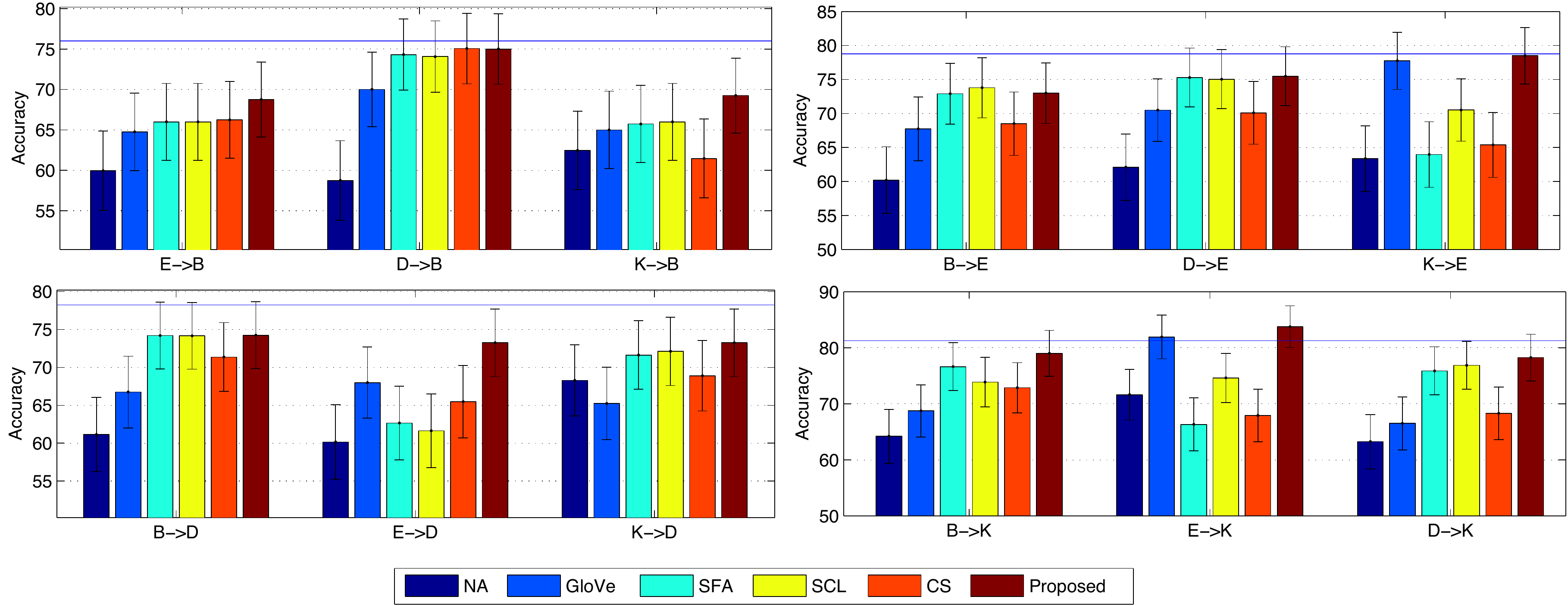}
\caption{Accuracies obtained by different methods for each source-target pair in cross-domain sentiment classification.}
\label{fig:overall}
\end{figure*}

In Figure~\ref{fig:overall}, we compare the proposed method against two baselines (\textbf{NA}, \textbf{InDomain}), 
current state-of-the-art methods for unsupervised cross-domain sentiment classification (\textbf{SFA}, \textbf{SCL}), 
word representation learning (\textbf{GloVe}), and cross-domain similarity prediction (\textbf{CS}).
The \textbf{NA} (no-adapt) lower baseline uses a classifier trained on source labeled data to classify target test data without any
domain adaptation. The \textbf{InDomain} baseline is trained using the labeled data for the target domain,
and simulates the performance we can expect to obtain if target domain labeled data were available.
Spectral Feature Alignment (\textbf{SFA})~\cite{Pan:WWW:2010} and
Structural Correspondence Learning (\textbf{SCL})~\cite{Blitzer:ACL:2007} are 
the state-of-the-art methods for cross-domain sentiment classification. However, those methods do not learn word representations. 

We use Global Vector Prediction (\textbf{GloVe})~\cite{Pennington:EMNLP:2014}, the current state-of-the-art word representation
learning method, to learn word representations separately from the source and target domain unlabeled data, and use the learnt representations
in Eq.~\ref{eq:score} for sentiment classification. In contrast to the \textit{joint} word representations learnt by the proposed method,
\textbf{GloVe} simulates the level of performance we would obtain by learning representations \textit{independently}.
\textbf{CS} denotes the cross-domain vector prediction method proposed by \newcite{Bollegala:ACL:2014}.
Although \textbf{CS} can be used to learn a vector-space translation matrix, it \emph{does not} learn word representations.
Vertical bars represent the classification accuracies (i.e. percentage of the correctly classified test instances)
obtained by a particular method on target domain's test data, and Clopper-Pearson $95\%$ binomial confidence intervals are superimposed. 

Differences in data pre-processing (tokenization/lemmatization), selection (train/test splits), feature representation (unigram/bigram),
pivot selection (MI/frequency), and the binary classification algorithms used to train the final classifier
make it difficult to directly compare results published in prior work. Therefore, we re-run the original algorithms
on the same processed dataset under the same conditions such that any differences reported in Figure~\ref{fig:overall}
can be directly attributable to the domain adaptation, or word-representation learning methods compared.

All methods use $l_2$ regularized logistic regression
as the binary sentiment classifier, and the regularization coefficients are set to their optimal values on the validation dataset.
\textbf{SFA}, \textbf{SCL}, and \textbf{CS} use the same set of $500$ pivots as used by the proposed method
 selected using NPMI (Section~\ref{sec:pivots}).
Dimensionality $n$ of the representation is set to $300$ for both \textbf{GloVe} and the proposed method. 

From Fig.~\ref{fig:overall} we see that the proposed method reports the highest classification accuracies in all $12$ domain pairs.
Overall, the improvements of the proposed method over \textbf{NA}, \textbf{GloVe}, and \textbf{CS} are statistically significant,
and is comparable with \textbf{SFA}, and \textbf{SCL}. The proposed method's improvement over \textbf{CS} shows the importance
of \emph{predicting} word representations instead of \emph{counting}. 
The improvement over \textbf{GloVe} shows that it is inadequate to simply apply existing word representation learning methods
 to learn independent word representations for the source and target domains. 
 
\begin{figure}[t]
\centering
\includegraphics[height=50mm]{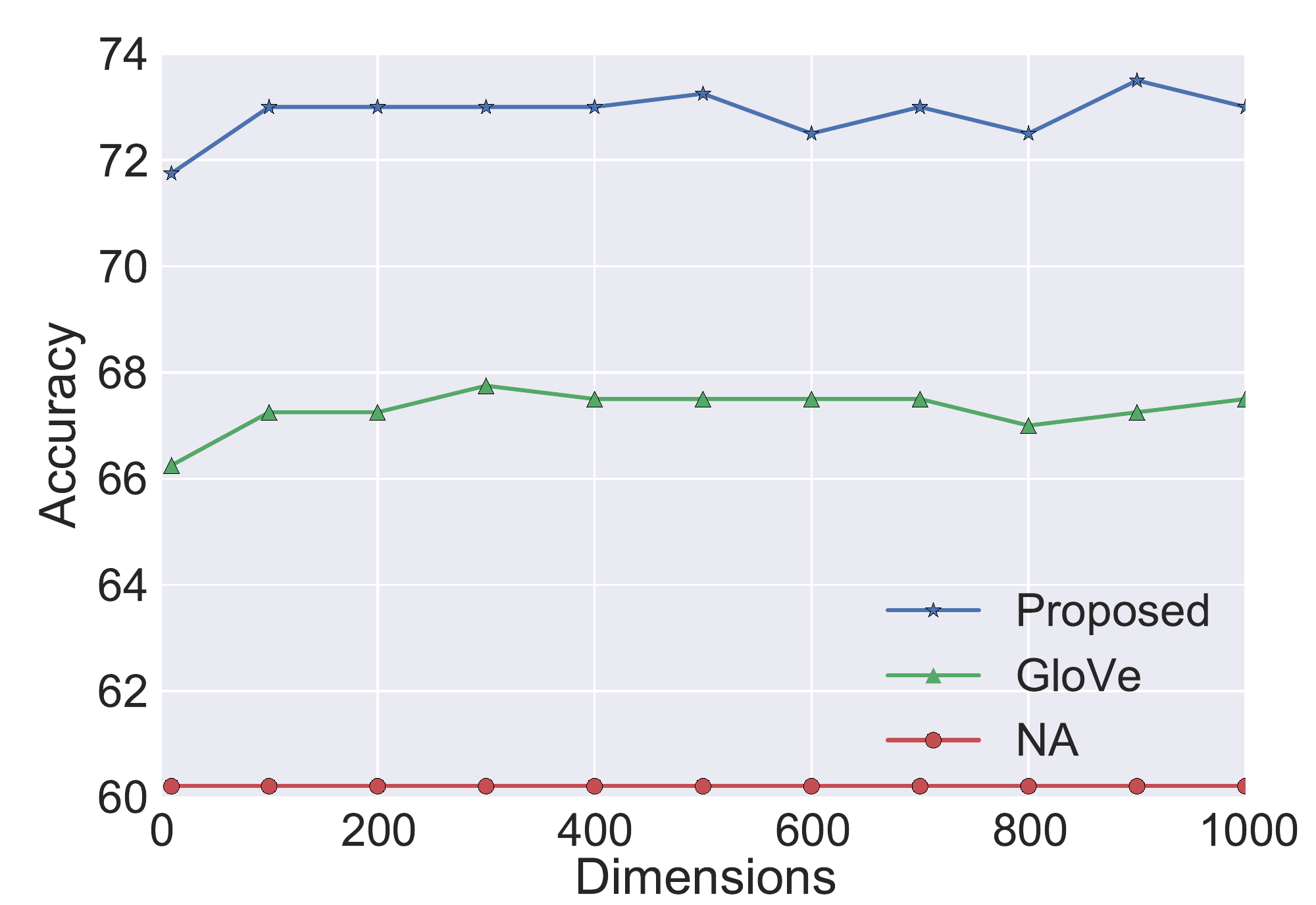}
\caption{Accuracy vs. dimensionality of the representation.}
\label{fig:dims}
\end{figure}

 We must consider the correspondences between the two domains as expressed by the pivots
 to jointly learn word representations. As shown in Fig.~\ref{fig:dims}, the proposed method reports superior accuracies over
\textbf{GloVe} across different dimensionalities. Moreover, we see that when the dimensionality of the representations increases,
initially accuracies increase in both methods and saturates after $200-600$ dimensions. However, further increasing the dimensionality
results in unstable and some what poor accuracies due to overfitting when training high-dimensional representations.
Although our word representations learnt by the proposed method are not specific to sentiment classification,
the fact that it clearly outperforms \textbf{SFA} and \textbf{SCL} in all domain pairs is encouraging, and implies the wider-applicability 
of the proposed method for domain adaptation tasks beyond sentiment classification.

\section{Conclusion}

We proposed an unsupervised method for learning cross-domain word representations using a given set of pivots
and non-pivots selected from a source and a target domain. Moreover, we proposed a domain adaptation 
method using the learnt word representations.

Experimental results on a cross-domain sentiment classification task showed that the proposed method outperforms several
competitive baselines and achieves best sentiment classification accuracies for all domain pairs.
In future, we plan to apply the proposed method to other types of domain adaptation tasks such as cross-domain part-of-speech
tagging, named entity recognition, and relation extraction.

Source code and pre-processed data etc. for this publication are publicly available\footnote{\url{www.csc.liv.ac.uk/~danushka/prj/darep}}.

\bibliographystyle{acl}
\bibliography{DARep}
\end{document}